\documentclass{article}
\usepackage{spconf,amsmath,graphicx}
\usepackage{amsfonts}

\newtheorem{theorem}{Theorem}[section]

\title{Inferring the Graph of Networked Dynamical Systems under Partial Observability and Spatially Colored Noise}
\name{ Augusto Santos\textsuperscript{\rm 1} \,\,\,\,\,\, Diogo Rente\textsuperscript{\rm 2}  \,\,\,\,\,\,    Rui Seabra\textsuperscript{\rm 2}  \,\,\,\,\,\,  José M. F. Moura\textsuperscript{\rm 2}\thanks{This work is partially supported by the U.S. National Science Foundation under Grant CCF 2327905. A. Santos was partially supported by FCT/MCTES  -- Foundation for Science and Technology, Portugal, -- through national funds and when applicable co-funded EU funds under the project UIDB/50008/2020 and project UIDP/00326/2020. José M. F. Moura was partially supported by the U.S. National Science Foundation under Grant CCN 1513936. D. Rente and R. Seabra were funded in part by FCT under the project UIDP/00326/2020. E-mails: augusto.santos@lx.it.pt; moura@andrew.cmu.edu; drobertd@andrew.cmu.edu, rvilarpo@andrew.cmu.edu.}}

\address{\textsuperscript{\rm 1} Instituto de Telecomunica\c{c}\~{o}es-IT, Lisbon, Portugal \\ \textsuperscript{\rm 2} Department of Electrical and Computer Engineering at Carnegie Mellon University, PA, USA}

\begin{document}
%
\maketitle
\begin{abstract}
In a Networked Dynamical System (NDS), each node is a system whose dynamics are coupled with the dynamics of neighboring nodes. The \emph{global} dynamics naturally builds on this network of couplings and it is often excited by a noise input with nontrivial structure. The underlying network is unknown in many applications and should be inferred from observed data. We assume: i) \emph{Partial observability}--- time series data is only available over a subset of the nodes; ii) \emph{Input noise}--- it is correlated across distinct nodes while temporally independent, i.e., it is \emph{spatially colored}. We present a \emph{feasibility condition} on the noise correlation structure wherein there exists a consistent network inference estimator to recover the underlying fundamental dependencies among the observed nodes. Further, we describe a structure identification algorithm that exhibits competitive performance across distinct regimes of network connectivity, observability, and noise correlation. 
\end{abstract}

\begin{keywords}
Network Inference, Complex Systems, Partial Observability, Colored Noise, Machine Learning.
\end{keywords}
%


\section{Introduction}

Complex Systems are defined by a set of interacting nodes whose connectivity structure determines several emergent patterns~\cite{dynamics_complex}, e.g., Brain activity~\cite{Brain3}; Gene Regulatory Networks~\cite{GRN3}; pandemics~\cite{topoepidemics}; or social networks~\cite{Jad}. Revealing the underlying network from observed data is fundamental in prediction, forecasting or control~\cite{Brain3,topoepidemics,Net_dismantling}. In these applications, only the time series data at some of the nodes are often available and the underlying network structure is unknown. The main goal of structure identification or network inference is to devise algorithms that yield consistent inference of the network structure from these observed data. 




A thorough discussion on structure identification related works can be found in~\cite{SMachado}. Several works focus on linear NDS, e.g.,~\cite{NIPS_Bento,Mei_latent,MaterassiSalapakaCDC2015,tomo_journal,open_Journal} and devise methods particularly tailored to sparse networks under diagonal noise, i.e., the excitation noise is assumed independent across nodes. The noise independence and network sparsity are central assumptions in Structural Causal Models (SCM)~\cite{pearl_2009}. These assumptions make network inference amenable to formal analysis, but they are not usually met in real world applications. On the other hand, the recent reference~\cite{Salapaka_Colored} considers a linear dynamic influence model (LDIM) perturbed by colored noise. It proposes a Wiener based approach tailored to sparse networks (leveraging on a sparse plus low-rank method) to recover the network connectivity. Reference~\cite{Salapaka_Colored} builds on two important assumptions: i) Noise is independent across latent nodes; ii) the noise correlations stem from affine interactions. Our work makes no such assumptions.

The problem is, in general, ill posed. It is well known, within the scope of high-dimensional statistical inference, that the feasibility of inference tasks is fundamentally contingent upon the distinct regimes of parameters or noise excitation levels~\cite{Barbier_noise,Barbier_fundamental}. In particular, the regime of parameters, observability or noise level determine whether information about the object of inference is carried by the observed data or fundamentally lost, rendering inference impossible in the latter case. Three main regimes are: $i)$ \emph{feasible}--- information is conveyed in the observed data and there is a known algorithm to retrieve it; ii) \emph{hard}--- information is conveyed, but there is no known consistent inference algorithm; iii) \emph{impossible}--- information is fundamentally lost.

Within the scope of linear NDS excited by colored noise and under partial observability, we describe a novel feasibility condition on the noise structure wherein structural information is in the observed time series data and can be consistently recovered via an algorithm. Further, building on our recently developed feature based causal inference paradigm~\cite{SMachado}, we describe an algorithm to recover the network with competitive performance across distinct regimes of connectivity, observability, and noise correlation. We assume no knowledge on the noise covariance. Reference~\cite{ASantos_2024} provides further details. 



\section{Problem Formulation} \label{sec:proposedresearch}

The time series data~$\left\{\mathbf{y}(n)\right\}_{n\in\mathbb{N}}$ is generated according to the following linear dynamical law 
\vspace*{-.25cm}
\begin{equation}\label{eq:model}
\mathbf{y}(n+1)=A\mathbf{y}(n)+\mathbf{x}(n+1),
\vspace{-5px}
\end{equation}
where~$\mathbf{y}(n)=\left[y_1(n)\,\,y_2(n)\,\,\ldots\,\,y_N(n)\right]^{\top}\in\mathbb{R}^N$ is the vector collecting the state~$y_i(n)$ of each node~$i=1,2,\ldots,N$ over time $n\in\mathbb{N}$; the noise process~$\mathbf{x}(n)\overset{d}\sim\mathcal{N}\left(0,\Sigma_x\right)$ is zero mean Gaussian with covariance~$\Sigma_x\in\mathbb{S}_+^N$; $\left\{\mathbf{x}(n)\right\}_{n\in\mathbb{N}}$ is independent across time~$n$; the noise $\mathbf{x}(n+1)$ is independent of $\mathbf{y}(n)$ for all $n$; the interaction matrix~$A\in \mathbb{R}_{+}^{N}$ is assumed nonnegative and symmetric; the spectral radius of~$A$ is assumed smaller than~$1$, $\rho(A)<1$, which renders the NDS~\eqref{eq:model} stable; and, finally, $A$ is stochastic modulo a multiplicative constant, i.e., $A=\rho \overline{A}$ where $\overline{A}$ is stochastic and $\rho<1$.

The main goal of structure identification is to infer the support of the submatrix $A_S$ underlying the network of interactions among the subset of observed nodes $S$, from the time series data stemming from the observed nodes. This can be done, first, by estimating the ground-truth matrix $A_S$ as 
\vspace{-5px}
\begin{equation}\nonumber
\widehat{A}_S^{(n)}= A_{S}+\mathcal{E}^{(n)}_S,
\vspace{-5px}
\end{equation}
where $\widehat{A}^{(n)}_S$ is an estimator from $n$ time series samples represented as the sum of the ground-truth interaction submatrix~$A_S$ and an error matrix~$\mathcal{E}^{(n)}_S$. For structural consistency of the estimator~$\widehat{A}^{(n)}_S$, we need to show that the entries of the error matrix~$\mathcal{E}^{(n)}_S$ do not vary too much, i.e., $\mathcal{E}^{(n)}_{\max}-\mathcal{E}^{(n)}_{\min}$ should be small enough, where $\mathcal{E}^{(n)}_{\max}$ and $\mathcal{E}^{(n)}_{\min}$ are the maximum and minimum entries of~$\mathcal{E}^{(n)}_S$. Formally, $\widehat{A}_S^{(n)}$ is structurally consistent whenever\footnote{While we need to ensure $\mathcal{E}_S^{(n)}\rightarrow 0$ for consistency, for \emph{structural consistency}, we only require ${\sf Osc}\left(\mathcal{E}^{(n)}_S\right)$ to be small enough w.h.p.}~\cite{ASantos_2024}
\vspace{-10px} 
\begin{equation}\label{ineq:separab}
{\sf Osc}\left(\mathcal{E}^{(n)}_S\right)\overset{\Delta}=\mathcal{E}^{(n)}_{\max}-\mathcal{E}^{(n)}_{\min}\leq \frac{A^{+}_{\min}}{2},
\vspace{-10px}
\end{equation}
where $A^{+}_{\min}$ is the smallest nonzero off-diagonal of $A_S$. In this case, the network can be recovered via properly thresholding the off-diagonals of $\widehat{A}^{(n)}_S$ (refer to Definition~$1$ in~\cite{SMachado}).

Therefore, two main steps are critical to establish feasibility or structural consistency of an estimator: $i)$ Characterizing the underlying error matrix~$\mathcal{E}^{(n)}_S$; $ii)$ estimating the variability of the off-diagonal entries of~$\mathcal{E}_S^{(n)}$. For example, these are done in~\cite{tomo_journal_proceedings,tomo_journal,open_Journal} for the Granger (and related matrix-valued estimators) under partial observability and diagonal noise. 

To offer a novel feasibility condition for the structure identification problem under partial observability and colored noise, in the next section, we resort to the $\widehat{R}_1(n)-\widehat{R}_3(n)$ estimator as reference to characterize its error and ascertain a condition where its off-diagonals variability is small enough, where we have defined $\widehat{R}_k(n)\overset{\Delta}=\frac{1}{n}\sum_{\ell=0}^{n-1} \mathbf{y}(\ell+k)\mathbf{y}(\ell)^{\top}$ as the $k^{\text{th}}$ lag empirical covariance matrix. Further, because this estimator is the linear combination of lag-moments, Lemma~$1$ in~\cite{SMachado} guarantees linear separability w.h.p. of certain features that we will use to train neural networks and recover the network.

\section{Feasibility under Structured Noise and Partial Observability}

Now, we show a characterization for the limiting error~$\mathcal{E}_S$ associated with the estimator~$\widehat{R}_1(n)-\widehat{R}_3(n)$, Theorem~\ref{th:representation}, and based on $\mathcal{E}_S$ off-diagonals variability, we establish a novel condition on the noise covariance $\Sigma_x$ to grant feasibility of the structure identification problem under partial observability and colored noise, Theorem~\ref{th:affine}. The estimator~$\widehat{R}_1(n)-\widehat{R}_3(n)$ was studied recently in~\cite{R1minusR3} under diagonal noise. Due to lack of space here, the proofs of the Theorems are in~\cite{ASantos_2024}.

We assume \emph{noise variance homogeneity} across nodes, i.e., $\sigma^2=\mathbb{E}\left[\mathbf{x}_i^2\right]$ for all $i$. In this case, we observe that $\sigma^2\geq \mathbb{E}\left[\mathbf{x}_i\mathbf{x}_j\right]$ $\forall{i,j}$, necessarily. 
Further, we assume that no two variables are (almost surely) equal, which implies (under the homogeneity assumption) strict inequality $\sigma^2>\mathbb{E}\left[\mathbf{x}_i \mathbf{x}_j\right]=\left[\Sigma_{x}\right]_{ij}$ for all $i\neq j$. In other words, the off-diagonals of~$\Sigma_x$ are strictly smaller than its diagonal.

Under these assumptions, we can uniquely cast $\Sigma_x$ as
\vspace{-5px}
\begin{equation}\label{eq:reprdecomp}\nonumber
\Sigma_x:=\sigma^2_{{\sf gap}}I+\beta\mathbf{1}\mathbf{1}^{\top}+\overline{\Sigma}, 
\vspace{-5px}
\end{equation}
where we have defined $\sigma^2_{{\sf gap}} \overset{\Delta}= \sigma^2-\max_{i\neq j} \mathbb{E}\left[\mathbf{x}_i\mathbf{x}_j\right]>0$ and $\beta$ stands for the average of the off-diagonal entries of $\Sigma_x$. We are ready to characterize the limiting error matrix $\mathcal{E}_S$ as a function of $\Sigma_x$ and the resulting feasibility condition.

\begin{theorem}[Error characterization]\label{th:representation}
We have
\begin{equation}\label{eq:partial}\nonumber
\frac{1}{{\sigma}^2_{\sf gap}}\left(\left[\widehat{R}_1(n)\right]_S-\left[\widehat{R}_3(n)\right]_S\right) \overset{n\rightarrow \infty}\longrightarrow A_S + \mathcal{E}_{S},
\end{equation}
where the limiting error matrix is given by
\begin{equation}\label{eq:mathcalE-1}\nonumber
\mathcal{E}_S\overset{\Delta}=\frac{1}{\sigma_{\sf gap}^2}\left[\beta\rho\mathbf{1}_S\mathbf{1}_S^{\top}+\left[\left(I-A^2\right)  \left(\sum_{i=0}^{\infty} A^{i+1} \overline{\Sigma} A^i\right)\right]_S\right],
\end{equation}
and the convergence holds in probability as the number of samples $n$ scales to infinite.
\end{theorem}

The error characterization in Theorem~\ref{th:representation} alongside the condition~\eqref{ineq:separab} imply the following feasibility condition.

\begin{figure*}[!t]
	\centering
\includegraphics[width=1.0\linewidth]{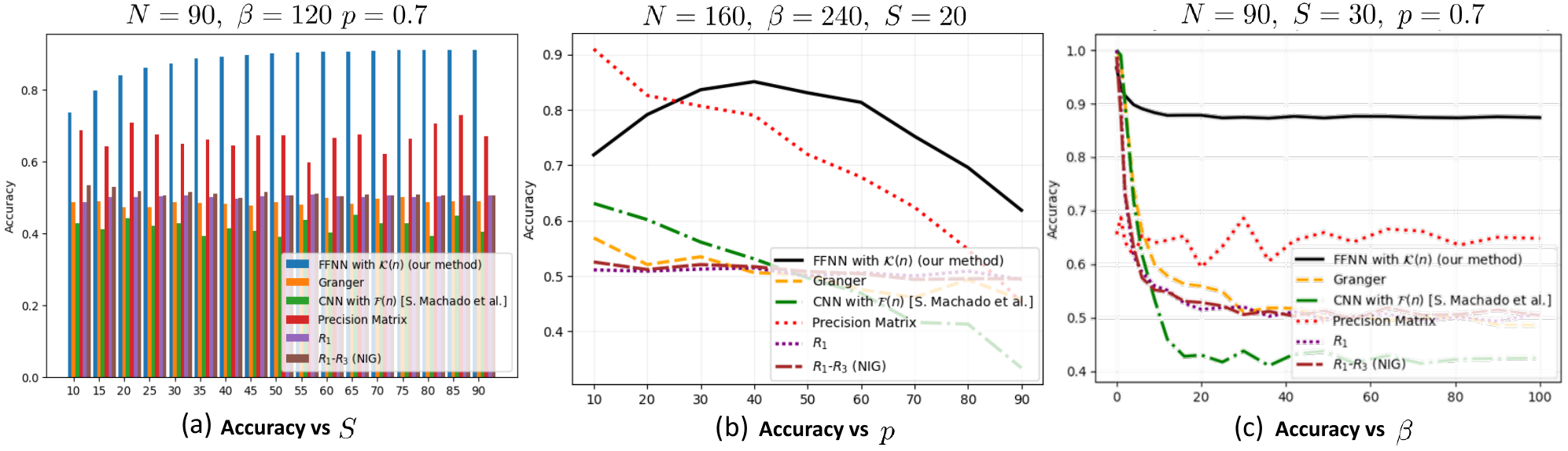}
	\caption{Plots (a)-(c) depict the accuracy of distinct estimators as a function of: (a) number of observed nodes $S$; (b) probability of connectivity~$p$ -- which controls the sparsity of the network; and (c) the offset parameter~$\beta$, respectively. The accuracy was computed with $n=5\times 10^5$ time series samples. For each plot, the remainder hyperparameters displayed at the top were frozen.}
	\label{fig:Performance}
	\vspace{-10px}
\end{figure*}

\begin{theorem}[Feasibility]\label{th:affine}
   If the covariance~$\Sigma_x$ obeys
    \begin{equation}\label{eq:exogeneous}
        \frac{{\sf Osc}\left({\sf Off}(\Sigma_x)\right)}{\sigma_{{\sf gap}}^2}\leq \frac{A^{+}_{\min}(1-\rho^2)}{2\rho(\rho^2+1)},
        \vspace{-5px}
    \end{equation}
    then $\left[\widehat{R}_1(n)\right]_S\!\!\!\!-\left[\widehat{R}_3(n)\right]_S$ is structurally consistent w.h.p., and ${\sf Off}\left(\Sigma_x\right)$ is the vector collecting the off-diagonals of $\Sigma_x$.
\end{theorem}

If we consider exogenous interventions in the NDS~\eqref{eq:model}, i.e., $\mathbf{y}(n+1)=A\mathbf{y}(n)+\mathbf{x}(n+1)+\mathbf{\xi}(n+1),$
where $\xi(n)\sim \mathcal{N}(0,\sigma_{\xi}^2 I)$ is i.i.d., and independent of the processes $\left(\mathbf{x}(n)\right)$ and $\left(\mathbf{y}(n)\right)$, then, we can control the variance-gap $\sigma_{\sf gap}^{2}$ in equation~\eqref{eq:exogeneous}. This implies, in particular, that we can recover structural consistency of $\left[\widehat{R}_1(n)\right]_S-\left[\widehat{R}_3(n)\right]_S$ regardless of the noise covariance~$\Sigma_x$, when the external excitation has large enough variance $\sigma^2_{\xi}$.  
\vspace{-10px}

\section{Feature Based Structure Identification Algorithm}
\vspace{-10px}

For structure identification, we adopt a feature based approach. Namely, we associate a feature vector to each pair of nodes in the NDS. These features are computed from the observed time series and shall abide by an important identifiability property: there exists one particular hyperplane that consistently partitions any set of features built upon the time series under distinct regimes of connectivity, observability and noise correlation. This property implies generalization of supervised methods trained with the features.    

Lemma $1$ in~\cite{SMachado} establishes that the structural consistency of the linear combination of lag-moments $\widehat{R}_1(n)-\widehat{R}_3(n)$ implies that the set of features~$\left\{\mathcal{F}_{ij}\right\}_{i\neq j}$ defined as 
\begin{equation}\label{eq:features}\nonumber
\mathcal{F}_{ij}(n)=\left(\left[\widehat{R}_D(n)\right]_{ij},\left[\widehat{R}_{D+1}(n)\right]_{ij},\ldots,\left[\widehat{R}_M(n)\right]_{ij}\right),
\end{equation}
with $D\leq 1$ and $M\geq 3$ are affinely separable. In view of the error representation in Theorem~\ref{th:representation}, we have that the average \emph{color} $\beta$ yields a \emph{drift away from the origin} of the features separating hyperplane while the matrix~$\overline{\Sigma}$ that describes the variation of the off-diagonals of~$\Sigma_x$ implies a spread of the features -- either compromising separability or rendering the problem \emph{hard}. To mitigate the drift (recovering stability of the separating hyperplane) and the spread, we use \emph{standard scaler}~\cite{scikit-learn} over the features $\left\{\mathcal{F}_{ij}(n)\right\}_{ij}$. 

Further, we observed empirically that the features given by the inverse of the observed part of the lag-moments  
\begin{equation}\label{eq:featurestau-1}\nonumber
 \mathcal{T}_{ij}(n) \!=\!       \left(\!\!\left[\!\!\left(\left[\widehat{R}_D(n)\right]_{S}\right)^{-1}\right]_{ij}\!\!,\ldots,\!\!\left[\left(\left[\widehat{R}_M(n)\right]_{S}\right)^{-1}\right]_{ij}\!\right),\nonumber
\end{equation}  
exhibited nontrivial identifiability properties under the framework considered, i.e., partial observability and colored noise, making them fit for structure identification. Taking the Cartesian product~$\mathcal{K}_{ij}(n)=\mathcal{F}_{ij}(n)\times \mathcal{T}_{ij}(n)$ resulted in features with greater separability under smaller number of samples.  

To generate each time series dataset~$\left\{\left[\mathbf{y}(\ell)\right]_S\right\}_{\ell}^{n}$, we first create a stable interaction matrix $A$ whose support reflects the underlying network of couplings. To this end, we proceed similarly to as done in~\cite{SMachado}, i.e., we draw a graph from an Erdős–Rényi random graph model. Then, we resort to the Laplacian rule~\cite{chungspectral} to weigh the generated graph so that the underlying matrix is stable (refer to~\cite{SMachado} for details).  
 
\textbf{Algorithm.} From the time series data~$\left\{\left[\mathbf{y}(\ell)\right]_S\right\}_{\ell=1}^{n}$, we compute the proposed features $\left\{\mathcal{K}_{ij}(n)\right\}$, where we have set $D=-50$ and $M=50$. To mitigate the features \emph{drift} caused by the offset $\beta$ in the noise covariance, we use \emph{standard scaler}~\cite{scikit-learn}. The normalized set of features~$\left\{\overline{\mathcal{K}}_{ij}(n)\right\}$ is used to train Feedforward Neural Networks (FFNN): the input of the neural network is given by the feature $\overline{\mathcal{K}}_{ij}(n)$ of the pair $ij$, and the output is given by the ground-truth, i.e., whether $ij$ is connected, $A_{ij}\neq 0$, or not. For training, we generate $11$ distinct time series datasets each one associated with a value of $\beta=0,5,10,\ldots,50$. All datasets build on a realization of an Erdős–Rényi random graph model with $N=50$ nodes and probability of connection $p=0.5$. As we will report next, our supervised method generalizes to other regimes of connectivity, observability and noise correlation.   


We will compare the performance of our algorithm with popular estimators for structure identification of linear NDS under partial observability: i) One lag $\left[\widehat{R}_1(n)\right]_S$ estimator; ii) the precision matrix $\left(\left[\widehat{R}_0(n)\right]_S\right)^{-1}$; iii) the NIG estimator $\widehat{R}_1(n)-\widehat{R}_3(n)$; iv) the Granger $\left[\widehat{R}_1(n)\right]_S\left(\left[\widehat{R}_0(n)\right]_S\right)^{-1}$; v) the feature based method proposed in~\cite{SMachado}. We deploy Gaussian mixture over the off-diagonals of the matrix estimators to classify the pairs as connected or disconnected. These are standard structural estimators known to resist the curse of observability and connectivity (dense or sparse networks).
\vspace{-10px}

\section{Numerical Experiments}
\vspace{-5px}

\begin{figure} [!t]
	\begin{center}	\includegraphics[scale= 0.32]{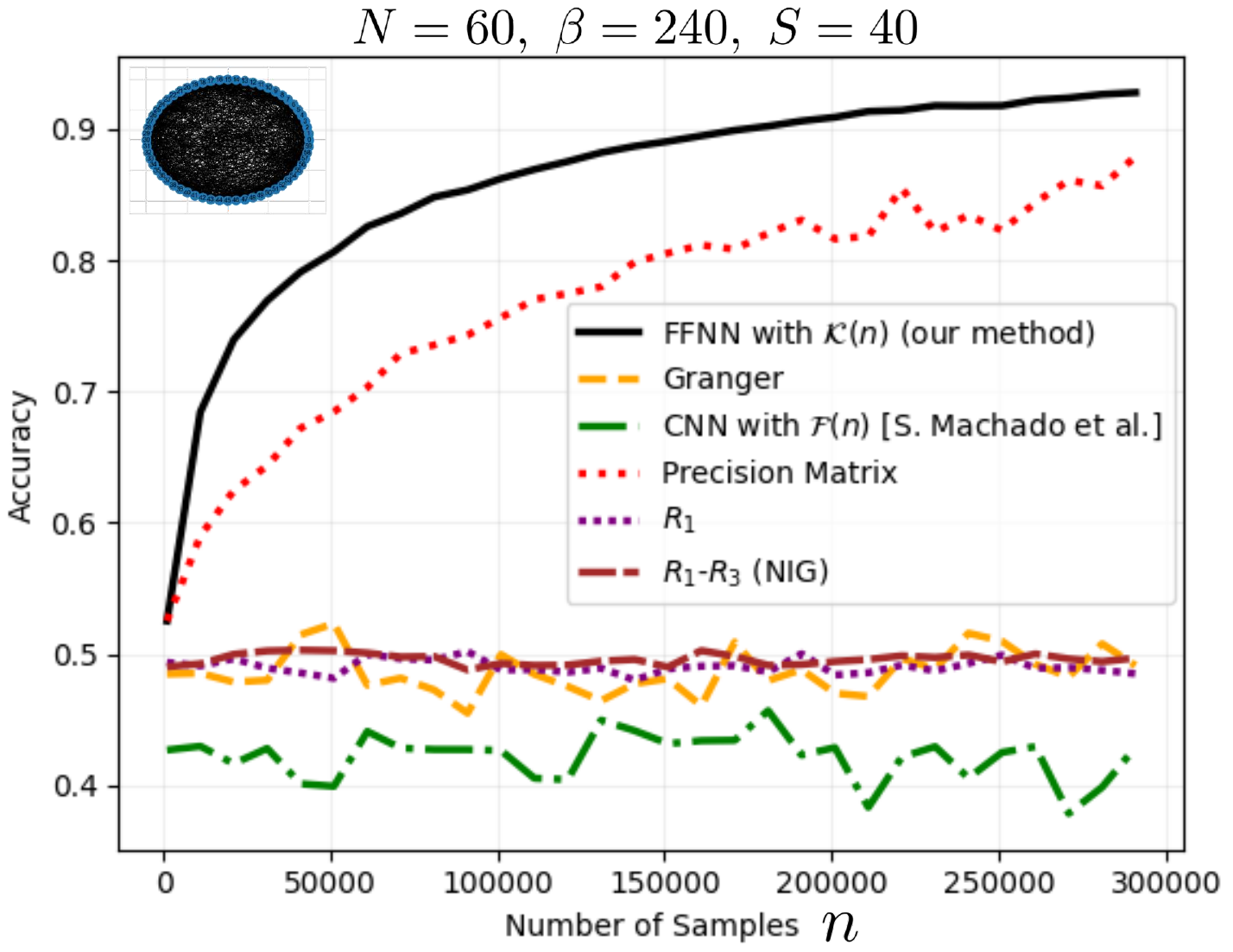}
		\caption{Synthetically generated small world network. In this network, about $67 \%$ of the pairs are connected.}
		\label{fig:Small}
	\end{center}
\vspace{-20px}
\end{figure}


First, we define \emph{accuracy} of the network recovery as the fraction of pairs correctly classified as connected or disconnected over all possible pairs of (observed) nodes in the NDS. 

Fig.~\ref{fig:Performance} entails the accuracy obtained by distinct estimators with $n=5\times 10^{5}$ time series samples as a function of: (a) the number of observed nodes $S$; (b) the probability of connection~$p$ between pairs of nodes under an Erdős–Rényi random graph model; (c) the offset parameter $\beta$ given by the average across the off-diagonals of $\Sigma_x$. The hyperparameters highlighted at the top of the plots represent the frozen parameters. For the experiments in Fig.~\ref{fig:Performance} (a) and (c), we have chosen a densely connected network drawn from an Erdős–Rényi random graph model with $p=0.7$. The dense regime is one of the most challenging settings within the scope of structure identification (under partial observability and colored noise).

Fig.~\ref{fig:Performance}(a) shows that our method withstands the curse of observability when contrasted with popular estimators including the feature based method~\cite{SMachado}. Fig.~\ref{fig:Performance}(b) illustrates that sparsity tends to favor the performance of the Precision matrix (or Graphical Lasso) whereas for \emph{denser} regimes, our method is overall better. The exact intersection point between the Precision and our method depends on the remainder hyperparameters. Fig.~\ref{fig:Performance}(c) shows that our method is able to display competitive accuracy across distinct regimes of noise correlation level. When $\beta=0$ the NDS is under the diagonal excitation noise. In this regime, the above estimators exhibit good performance with technical guarantees. For example, Granger and $R_1$ are structurally consistent~\cite{open_Journal}; The NIG $\widehat{R}_1(n)-\widehat{R}_3(n)$ was studied in~\cite{R1minusR3}; and the features in~\cite{SMachado} can be partitioned by a particular hyperplane with an overall competitive performance. As $\beta$ increases, the results in the diagonal noise domain collapse: i) Granger and $\widehat{R}_1(n)$ loose technical guarantees; ii) The NIG $\widehat{R}_1(n)-\widehat{R}_3(n)$ estimator creates a shift and perturbs the entries of the ground-truth $A_S$ in view of our Theorem~\ref{th:representation}; iii) the separating hyperplane of the features in~\cite{SMachado} \emph{drifts away} from the origin, penalizing the normalization done in~\cite{SMachado} and the resulting performance. 

The accuracy of our method displayed in Fig.~\ref{fig:Performance} across distinct regimes of observability, connectivity and noise correlation, further highlights that while the FFNNs were trained under full observability and over Erdős–Rényi generated graphs with $N=50$ nodes and probability $p=0.5$ of edge placing between nodes, they generalize well for partial observability, distinct connectivities, and noise correlation levels. 

\begin{figure} [!t]
	\begin{center}	\includegraphics[scale= 0.33]{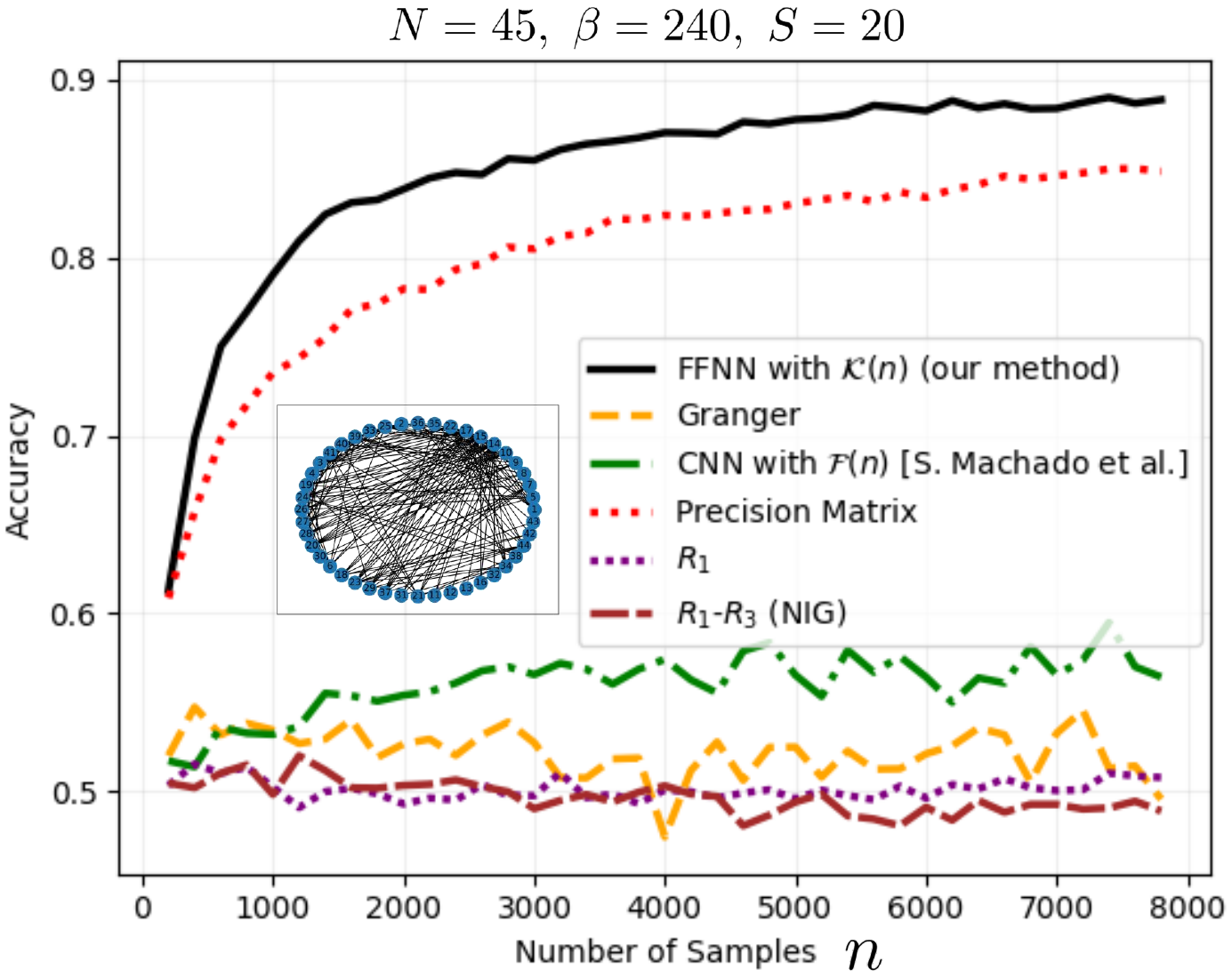}
		\caption{Real small world network: Caribbean Food web~\cite{foodweb}. In this network, about $20\%$ of the pairs are connected.}
		\label{fig:Small2}
	\end{center}
\vspace{-20px}
\end{figure}

To further highlight the generalization of our supervised method, in Fig.~\ref{fig:Small} and \ref{fig:Small2}, we depict the sample complexity performance of the several estimators over a synthetically generated small world network (dense network) and a real small world network (sparse network) underlying a Caribbean food web~\cite{foodweb}. While our supervised method was trained on Erdős–Rényi graphs, it displays good performance over structurally distinct graphs, like small world ones. 
\vspace{-10px}

\section{Concluding Remarks}
\vspace{-5px}

We considered the problem of network identification from partially observed times series streaming from a linear NDS excited by colored noise. First, we reported a novel condition on the noise structure to guarantee feasibility of the inference problem under partial observability. Then, leveraging our recent feature based causal inference work~\cite{SMachado}, we considered novel features to describe a structure identification algorithm with competitive performance across distinct regimes of connectivity, observability, and noise correlation as shown in the numerical results. Proofs of the results here are in~\cite{ASantos_2024}.




\bibliographystyle{IEEEbib}
\bibliography{biblio}

\end{document}